\documentclass{article}
\usepackage{iclr2019_conference,times}

\pdfoutput=1

\usepackage[utf8]{inputenc} 
\usepackage[T1]{fontenc}    
\usepackage[hidelinks]{hyperref}
\usepackage{url}            
\usepackage{booktabs}       
\usepackage{amsfonts}       
\usepackage{nicefrac}       
\usepackage{microtype}      
\usepackage[title]{appendix}
\usepackage{tikz}
\usepackage{graphicx}
\usepackage{booktabs}
\usepackage{multirow}
\usepackage{enumitem}
\usepackage{xspace}
\usepackage{booktabs}
\usepackage{float}
\usepackage{color}
\usepackage{amsmath}
\usepackage{amssymb}
\usepackage{xcolor}
\usepackage[normalem]{ulem}

\usepackage[font=small]{caption}
\usepackage{subcaption}

\usepackage{pgfplots,tikz}
\pgfplotsset{compat=1.9}

\bibpunct{(}{)}{;}{a}{}{,}

\DeclareMathOperator*{\argmax}{arg\,max}
\newcommand{\Loss}{\mathcal{L}}
\newcommand{\N}{\mathcal{N}}
\newcommand{\I}{\mathrm{I}}
\newcommand{\E}{\mathop{{}\mathbb{E}}}
\DeclareMathOperator{\KL}{KL}
\DeclareMathOperator{\var}{var}
\newcommand{\w}{\omega}

\newcommand{\ptb}{PTB\xspace}
\newcommand{\nlltoppl}[1]{%
  \pgfmathparse{exp(#1)}%
  \pgfmathprintnumber[fixed,zerofill,precision=1,assume math mode=true]{\pgfmathresult}}
\newcommand{\nlltopplbold}[1]{%
  \pgfmathparse{exp(#1)}%
  \textbf{\pgfmathprintnumber[fixed,zerofill,precision=1,assume math mode=true]{\pgfmathresult}}}

\definecolor{darkbrown}{RGB}{128, 64, 8}
\definecolor{darkblue}{RGB}{0, 0, 83}

\allowdisplaybreaks

\title{Pushing the bounds of dropout}

\author{Gábor Melis,~~~~Charles Blundell,~~~~Tomá\v{s} Ko\v{c}isk\'y,\\\textbf{Karl Moritz Hermann,~~~Chris Dyer,~~~Phil Blunsom}\\
  DeepMind, London, UK\\
  \texttt{\{melisgl,cblundell,tkocisky,kmh,cdyer,pblunsom\}@google.com}\\
}

\iclrfinalcopy 
\begin{document}

\maketitle

\begin{abstract}
  We show that dropout training is best understood as performing MAP
  estimation concurrently for a family of conditional models whose
  objectives are themselves lower bounded by the original dropout
  objective.
  This discovery allows us to pick any model from this family after
  training, which leads to a substantial improvement on
  regularisation-heavy language modelling.
  The family includes models that compute a power mean over the
  sampled dropout masks, and their less stochastic subvariants with
  tighter and higher lower bounds than the fully stochastic dropout
  objective.
  We argue that since the deterministic subvariant's bound is equal to
  its objective, and the highest amongst these models, the predominant
  view of it as a good approximation to MC averaging is misleading.
  Rather, deterministic dropout is the best available approximation to
  the true objective.
\end{abstract}

\section{Introduction}

The regularisation technique known as dropout underpins numerous
state-of-the-art results in deep learning
\citep{hinton2012improving,srivastava2014dropout}, and its application
has received much attention in the form of optimisation
\citep{wang2013fast} and attempts at explaining or improving its
approximation properties
\citep{baldi2013understanding,zolna2017fraternal,DBLP:journals/corr/MaGHYDH16}.
The dominant perspective today views dropout as either an implicit
ensemble method \citep{warde2013empirical} or averaging over an
approximate Bayesian posterior \citep{gal2016dropout}.
Regardless of which view we take, dropout training is carried out the
same way, by minimising the expectation of the loss over randomly
sampled dropout masks.
However, at test time these views naturally lead to different
algorithms: the Bayesian approach computes an arithmetic average as it
marginalises out the weight uncertainty, while the ensemble approach
typically uses the geometric average due to its close relationship to
the loss.
Collectively they are called MC dropout and neither is clearly better
than the other \citep{warde2013empirical}.
A third way to make predictions is to ``turn dropout off'', that is,
propagate expected values through the network in a single,
deterministic pass.
This deterministic (also known as \emph{standard}) dropout in
considered to be an excellent approximation to MC dropout.

This situation is unsatisfactory as it does not provide theoretical
grounding for dropout, without which the choice of dropout variant
remains arbitrary.
In this paper, we provide such theoretical foundations.
First, we prove the dropout objective to be a common lower bound on
the objectives of a family of infinitely many models.
This family includes models corresponding to the three aforementioned
methods of evaluation: the arithmetic averaging, the geometric
averaging, and the deterministic.
Thus by maximising the dropout objective we get a single set of
parameters and many models that all have the same parameters but
differ in how they make predictions.
This allows us to train once and perform model selection at validation
time by evaluating the different methods of making predictions
corresponding to individual models in the family.
Second, we turn the conventional perspective on its head by showing
that while dropout training performs stochastic regularisation, the
trained model is best viewed as deterministic, not as a stochastic
model with a deterministic approximation.

This paper is structured as follows.
In \S\ref{sec:variational-dropout}, we revisit variational dropout
\citep{gal2016dropout} and demonstrate that, despite common
perception, sharing of masks is not necessary, neither in theory nor
in practice.
Then, by recasting dropout in a simple conditional form, we highlight
the counterintuitive role played by the variational posterior.
\S\ref{sec:the-dropout-family} contains our main contributions.
Here we construct a family of conditional models whose MAP objectives
are all lower bounded by the usual dropout objective, and identify a
member of this family as best in terms of model fit.
In \S\ref{sec:applying-dropout}, we select the best of this family in
terms of generalisation to improve language modelling.
Finally, creating a cheap approximation to the bias of this model
allows us to get better results from model tuning.

\section{Variational dropout}
\label{sec:variational-dropout}

Since its original publication \citep{hinton2012improving}, dropout
had been considered a stochastic regularisation method, implemented as
a tweak to the loss function.
That was until \mbox{\cite{gal2016dropout}} grounded dropout in
much-needed theory.
Their subsequent work \citep{gal2016theoretically} focused on RNNs,
showing that if dropout masks are shared between time steps, the
objective for their proposed variational model is the same as the
commonly used dropout objective with an $\ell_2$ penalty.
Their method became known as \textbf{variational dropout}, not to be
confused with \cite{kingma2015variational}, and is used in
state-of-the-art sequential models
\citep{merity2017regularizing,melis2017state}.
Before we move on to a more general formulation we revisit it to
better understand its critical features.

First, we recall the derivation of variational dropout.
Consider an RNN that takes input $x$ and maps it to output $y$ and is
trained on a set of $N$ data points in paired sets $X$, $Y$.
A variational lower bound on the log likelihood is obtained as
follows:
\begin{align}
 \ln p(Y|X)
 &= \ln \E_{\w \sim q(\w)} \frac{p(Y|X,\w)p(\w)}{q(\w)} \nonumber\\
 &\geqslant \E_{\w \sim q(\w)} \ln p(Y|X,\w) - \KL(q(\w)||p(\w)) \nonumber\\
 &= \int q(\w) \ln p(Y|X,\w)d\w - \KL(q(\w)||p(\w)) \nonumber \\ 
 \label{eq:loss}
 &= \sum_{i=1}^N \int q(\w) \ln p(y_i|x_i,\w)d\w - \KL(q(\w)||p(\w)),
\end{align}
where $p(y|x,\w)$ is defined by the RNN with weights $\w$. Variational
Bayesian methods then maximise this lower bound with respect to the
variational distribution $q(\w)$.
For variational dropout, $q(\w)$ takes the form of a mixture of two
gaussians with small variances: one with zero mean that represents the
dropped out rows of weights, and another with mean $\Theta$:
\begin{align*}
    q(\w_r) &= p \N(\w_r|0, \sigma^2\I) + (1-p) \N(\w_r| \Theta_r, \sigma^2\I)
\end{align*}
In the above, $r$ is the index of a row of a weight matrix.
Dropping whole rows of weights is equivalent to the more familiar view
of dropout over units.
The prior over the weights is a zero mean gaussian:
\begin{align*}
    p(\w) &= \N(\w|0, \sigma_p^2\I)
\end{align*}
The loss is defined based on Eq.\,\ref{eq:loss}.
The integrals are approximated using a single sample $\hat{\w} \sim
q(\w)$, and the KL term is approximated with weight decay on $\Theta$:
\begin{align}
    \label{eq:approximate-loss}
    \Loss = - \sum_{i=1}^N \ln p(y_i|x_i,\hat{\w}_i) + \KL(q(\w)||p(\w))
\end{align}

The same dropout mask (and consequently the same $\w$) is employed at
every time step.
This sharing of masks is considered the defining characteristic of
variational dropout, but we note in passing that the theory for the
non-shared masks case is very similar and there is little between them
in practice with LSTMs (see Appendix\,\ref{sec:non-shared-mask}).
With this we conclude the recap of variational dropout, and describe
our contributions in the rest of the paper.

\subsection{Dropout as a conditional model}
\label{sec:arithmetic}

In variational inference the idea is to approximate the intractable
and complicated posterior with a simple, parameterised distribution
$q$.
Crucially, this approximation affects our inferences and predictions.
If we are serious about it being an approximation to the posterior
and want to reduce its distortion of the model $p$, then $q$ can be
made more flexible.
But making $q$ more flexible in variational dropout can potentially
ruin the regularisation effect.
So the particular choice of $q$ plays an important, active role: it
effectively performs posterior regularisation and acts as an integral
part of the model.

Coming from another angle, \cite{Osband2016RiskVU} makes the point
that in variational dropout the posterior over weights does not
concentrate with more data, unlike for example in
\cite{graves2011practical}, which is unexpected behaviour from a
Bayesian model.
This conundrum is caused by encoding dropout with a fixed rate mixture
of fixed variance components in $q$,
which also necessitates expensive tuning of the dropout rate.
\cite{gal2017concrete} proposes a way to address these shortcomings.

To avoid getting bogged down in the issues surrounding the suitability
of variational inference and ease interpretation, we construct a
straightforward conditional model and lower bound its MAP objective in
the same form as the variational objective.
Suppose we want to do MAP estimation for the model parameters (the
means of the distribution of weights, $\Theta$): $\argmax_{\Theta}
p(\Theta|X,Y)$.
Consider a conditional model $p(Y|X,\Theta)$ as a crippled generative
model with $p(x_i)$ constant, $x_i$ and $\Theta$ independent.
Place a normal prior on the means $\Theta$ and otherwise make the
weights $\w$ conditional on $\Theta$ the same way as they were in the
variational posterior $q(\w)$:
\begin{align}
  p(\Theta) &= \N(\Theta|0, \sigma_p^2\I) \nonumber\\
  p(\w_r|\Theta) &= p \N(\w_r|0, \sigma^2\I) +
  (1-p) \N(\w_r| \Theta_r, \sigma^2\I) \nonumber\\
  \label{eq:arithmetic-model}
  p(y,\w|x,\Theta) &= p(y|x,\w) p(\w|\Theta)
\end{align}
The log posterior of this model has a similar lower bound to the
variational objective (Eq.\,\ref{eq:loss}):
\begin{align}
  \label{eq:map-objective}
  \ln p(\Theta|X,Y) &\geqslant \sum_{i=1}^N \int p(\w|\Theta) \ln p(y_i|x_i,\w) d\w + \ln p(\Theta) - C_{\textit{MAP}}
\end{align}
See Appendix\,\ref{sec:map-objective-derivation} for detailed
derivation.
Dropping the normalisation constant $C_{\textit{MAP}}$ that doesn't
depend on $\Theta$, and approximating the above integrals with a
single sample, the loss corresponding to the MAP objective becomes:
\begin{align}
    \label{eq:approximate-map-loss}
    \Loss_{\textit{MAP}} &= -\sum_{i=1}^N \ln p(y_i|x_i,\hat{\w}_i) - \ln p(\Theta)
\end{align}
The first term of this loss is identical to that of the loss for
variational dropout (Eq\,\ref{eq:approximate-loss}).
If the prior on $\Theta$ is a zero mean gaussian, then the second term
is equivalent to a weight decay penalty just like the KL term in the
variational setup.
With the two losses being effectively the same, in the following we
focus on MAP estimation for the conditional model to sidestep any
questions about whether variational inference makes sense in this
case.

\section{The dropout family of models}
\label{sec:the-dropout-family}

Having developed a conditional model for dropout that leads to the
same objective as variational dropout, we now derive a family of
models whose objectives are all lower bounded by the usual dropout
objective.
We draw inspiration from the different evaluation methods employed for
dropout:
\begin{itemize}
\item \emph{Deterministic dropout} propagates the expectation of each
  unit through the network in single pass. This is very efficient and
  is viewed as a good approximation to the next option.
\item \emph{MC dropout} mimicks the training procedure, and averages
  the predicted probabilities over randomly sampled dropout masks.
  With one forward pass per sample, this can be rather expensive.
  There is some ambiguity as to what kind of averaging shall be
  applied: oftentimes the \emph{geometric} average (GMC) is used,
  because of its close relationship to the loss, but the
  \emph{arithmetic} average (AMC) is also widespread.
\end{itemize}

Our goal in this section is to demonstrate the consequences of
optimising a lower bound instead of the true objective.
While it is easy to argue in general that objectives of more than one
model may share any given lower bound, for dropout a particularly
simple explicit construction of such a family of models is possible.
As we will see, this allows for post-training model selection based on
validation results given a trained set of parameters.
In the absence of validation results to guide model selection,
inspection of the tightness of the lower bound indicates the
deterministic model as the most reasonable choice from the family.

\subsection{Geometric model}
\label{sec:geometric}

First, we investigate whether the geometric or the arithmetic mean is
the correct choice for making predictions in the context of
classification.
Recall the predictive term of the MAP loss in
Eq.\,\ref{eq:approximate-map-loss}: $\sum \ln p(y_i|x_i,\hat{\w}_i)$.
Notice how with SGD and multiple epochs, for each data point several
dropout masks are encountered, and the approximating quantity becomes
the geometric mean of the predicted probabilities $p(y_i|x_i,\w)$ over
the masked weights.
For this reason, the posterior predictive distribution
$p(y^*|x^*,X,Y)$ is often computed as the renormalised geometric mean.
This is in apparent conflict with the conditional model that
prescribes the arithmetic mean (integrating $\w$ out of
Eq.\,\ref{eq:arithmetic-model}).
However, we can define another model where the conditional
distribution is directly defined to be the renormalised geometric mean
\begin{align}
  \label{eq:renormalised-geometric-model}
  p(y|x,\Theta) &= \frac{\exp\!\big(\E_{\hat{\w} \sim p(\w|\Theta)} \ln p(y|x,\hat{\w})\big)}{Z(x,\Theta)}, &
   Z(x,\Theta) &= \sum_{c=1}^C \exp\big(\E_{\hat{\w} \sim p(\w|\Theta)} \ln p(c|x,\hat{\w})\big)
\end{align}
with a slight abuse of notation, due to using the symbol $p$ in
$p(y|x,\Theta)$ although $p(y|x,\Theta) \neq \E_{\w} p(\w|\Theta)
p(y|x,\w)$.
It can be shown that the arithmetic model's (Eq.
\ref{eq:arithmetic-model}) lower bound (Eq. \ref{eq:map-objective}) is
a lower bound for this renormalised geometric model (Eq.
\ref{eq:renormalised-geometric-model}), as well.
See Appendix\,\ref{sec:geometric-objective-derivation} for the
derivation.
The answer to the question whether we should use GMC or AMC is that it
depends: they correspond to different models, but the dropout
objective is a lower bound on the objectives of both models.
So one can freely choose between GMC and AMC \emph{at evaluation time},
doing model selection retrospectively after training.

\subsection{The power mean model family}
\label{sec:power-mean-family}

Having two models to choose from, it is natural to ask whether these
are just instantiations of a larger class of models.
We propose the power mean family of models to extend the set of models
to a continuum between the geometric and arithmetic models described
in §\ref{sec:geometric} and §\ref{sec:arithmetic}, respectively, and
show that they have the same lower bound.
The power mean is defined as:
\begin{align*}
M_\alpha(x_1,\dots,x_n) = \bigg(\frac{1}{n}\sum_{i=1}^n x_i^\alpha\bigg)^{1/\alpha}
\end{align*}
For $\alpha=1$ we arrive at the arithmetic mean while the natural
extension to $\alpha=0$ is the geometric mean as it is the limit of
$M_\alpha$ at $\alpha \to 0$, which can be proven with L'Hôpital's
rule.
Similarly to the construction of the geometric model, we define the
power mean model by directly conditioning on $\Theta$:
\begin{align}
  \label{eq:power-mean-model}
  p(y|x,\Theta) &= \frac{\sqrt[\alpha]{\E_{\hat{\w} \sim p(\w|\Theta)} p(y|x,\hat{\w})^\alpha}}{Z(x,\Theta)}, &
  Z(x,\Theta) &= \sum_{c=1}^C \sqrt[\alpha]{\E_{\hat{\w} \sim p(\w|\Theta)} p(c|x,\hat{\w})^\alpha}
\end{align}
where $Z(x,\Theta)$ is at most $1$ if $\alpha \in (-\infty,1]$ because
  $M_{\alpha}$ is monotonically increasing in $\alpha$ and $Z$ is $1$
  for $\alpha=1$.
Here we provide a concise derivation of a lower bound on the log
posterior (the full derivation can be found in
Appendix\,\ref{sec:power-mean-objective-derivation}):
\begin{align}
  \ln p(\Theta|X,Y) &= \sum_{i=1}^N \bigg[\ln \sqrt[\alpha]{\E_{\hat{\w} \sim p(\w|\Theta)} p(y_i|x_i,\hat{\w})^\alpha} - \ln(Z(x_i,\Theta))\bigg] + \ln p(\Theta) - C_{\textit{MAP}} \nonumber\\
  \label{eq:power-mean-z}
  &\geqslant \sum_{i=1}^N \ln \sqrt[\alpha]{\E_{\hat{\w} \sim p(\w|\Theta)} p(y_i|x_i,\hat{\w})^\alpha} + \ln p(\Theta) - C_{\textit{MAP}}\\
  \label{eq:power-mean-jensen}
  &\geqslant \sum_{i=1}^N \frac{1}{\alpha} \E_{\hat{\w} \sim p(\w|\Theta)} \ln p(y_i|x_i,\hat{\w})^\alpha + \ln p(\Theta) - C_{\textit{MAP}}\\
  &= \sum_{i=1}^N \int p(\w|\Theta) \ln p(y_i|x_i,\w) d\w + \ln p(\Theta) - C_{\textit{MAP}}\nonumber
\end{align}
The first inequality above follows from $Z(x,\Theta)\leqslant 1$ for
all $x$, $\Theta$, while the second is an application of Jensen's rule
assuming $\alpha>0$.
We arrived at the same lower bound on the objective as we had for the
geometric (Eq.\,\ref{eq:renormalised-geometric-model}) and arithmetic
models (Eq. \ref{eq:arithmetic-model}), thus defining the
\textbf{power mean family} with parameter $\alpha\in[0,1]$ of models
from which we can choose at evaluation time. For $\alpha>1$, the
normalising constant $Z$ would be greater than $1$, and this would not
be a lower bound in general.

\subsection{Tightness of the lower bound}

To better understand the quality of fit for models in the power mean
family we examine the tightness of their lower bounds.
There are two steps involving inequalities in the derivation of the
bound: one where the normalisation constant $Z$ is dropped
(Eq.\,\ref{eq:power-mean-z}) and another where the logarithm is moved
inside the expectation (Eq. \ref{eq:power-mean-jensen}).
We show that the gaps introduced by these steps can be made
arbitrarily small by reducing the variance of $p(y|x,\w)$ with respect
to $\w$.

Notice that the Jensen gap with the logarithm function is scale
invariant:
\begin{align*}
\ln(\E[\lambda L]) - \E\ln(\lambda L) = \ln(\E L) - \E\ln(L)
\end{align*}
Intuitively, this suggests that $\var(L) / (\E L)^2$ is closely
related to the size of the gap.
Indeed, \cite{maddison2017filtering} show that if the first inverse
moment of $L$ is finite, then
\begin{align}
\label{eq:jensen-gap-approximation}
\ln(\E L) - \E\ln(L)) =
\frac{\var(L)}{2(\E L)^2} + \mathcal{O}(\sqrt{\E[(L - \E L)^6]})
\end{align}
Here we go a bit further and show that if there is a positive lower
and upper bound on $L$, then there are non-trivial lower and upper
bounds on its Jensen gap and these are bounds are multiplicative in
$\var(L)$.
Let $L$ be a random variable such that $P(L\in(a,b)) = 1$ where
$-\infty \leqslant a < b \leqslant \infty$. Furthermore, let
$\varphi(l)$ be a convex function. Jensen's inequality states that
$E[\varphi(L)] \geqslant \varphi(\E[L])$. \cite{liao2017sharpening}
show that the Jensen gap $E[\varphi(L)] - \varphi(\E[L])$ can be
bounded from below and above:
\begin{gather}
\label{eq:bound-bound}
\inf\{h(l;\mu) \mid l \in (a,b)\} \var(L) \leqslant \E[\varphi(L)] - \varphi(\E[L]) \leqslant \sup\{h(l;\mu) \mid l \in (a,b)\} \var(L)\nonumber\\
h(l;\mu) = \frac{\varphi(l) - \varphi(\mu)}{(l-\mu)^2} - \frac{\varphi'(\mu)}{l-\mu} \nonumber
\end{gather}
where $h(l; \mu)$ does not depend on the distribution of $L$, only on
its expected value $\mu$ and on the function~$\varphi$.
Substituting $L=p(c|x,\w)$ (a random variable on $[0,1]$ due to the
randomness of the dropout masks) and $\varphi(l)=-\ln(l)$, we can see
that the gap introduced by Eq.\,\ref{eq:power-mean-jensen} can be made
smaller by decreasing the variance of the predictions while
maintaining the expected value of $L$ (i.e. the expected probability),
assuming that there is a positive lower and upper bound on them (so
that the supremum is finite and the infimum is positive,
respectively).
A similar argument based on $\sum_{c=1}^C M_\alpha( \E_{\hat{\w}}
p(c|x,\hat{\w})) = 1$ shows that $Z$ approximately monotonically
approaches $1$ as the variance decreases, so the gap of
Eq.\,\ref{eq:power-mean-z} can also be reduced.

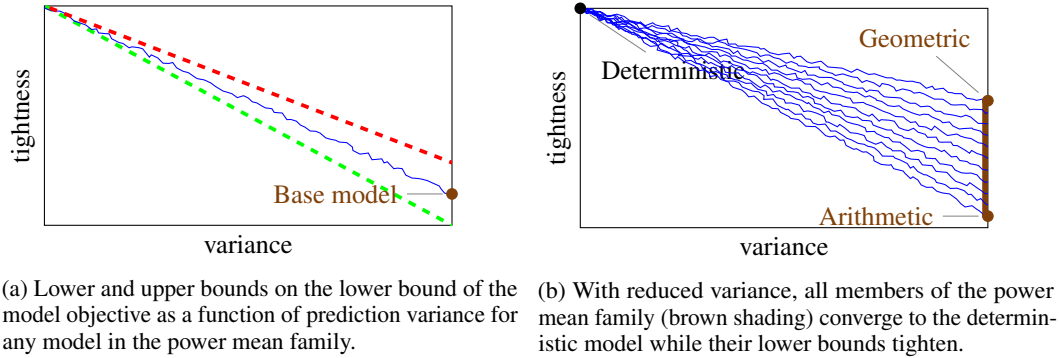
\begin{figure}[!t]
  \begin{subfigure}{0.49\linewidth}
    \begin{tikzpicture}
      \begin{axis}[xmin=0,xmax=1.0,ymin=0.3,ymax=1, samples=50,
          width=7cm,height=4.5cm,
          xlabel=variance,ylabel=tightness,ticks=none]
          \addplot[green, dashed, line width=0.05cm] (x, 1-0.7*x);
          \addplot[blue, decorate, decoration={random steps,segment length=2pt,amplitude=1pt}] (x, 1-0.6*x);
          \addplot[red, dashed, line width=0.05cm] (x, 1-0.5*x);
          \addplot[darkbrown, mark=*] coordinates {(1.0,0.4)} node[pin=180:{Base model}]{} ;
      \end{axis}
    \end{tikzpicture}
    \caption{\small Lower and upper bounds on the lower bound of the
      model objective as a function of prediction variance for any
      model in the power mean family.}
    \label{fig:lower-bound-bounds}
  \end{subfigure}
  \hfill
  \begin{subfigure}{0.49\linewidth}
    \begin{tikzpicture}
      \begin{axis}[xmin=0,xmax=1,ymin=0.05,ymax=1, samples=50,
          width=7cm,height=4.5cm,
          xlabel=variance,ylabel=tightness,ticks=none]
        \addplot[mark=*] coordinates {(0,1)} node[pin=-75:{Deterministic}]{} ;
        \addplot[darkbrown, mark=*] coordinates {(1,0.6)} node[pin=100:{Geometric}]{} ;
        \addplot[darkbrown, mark=*] coordinates {(1.0,0.1)} node[pin=180:{Arithmetic}]{} ;
        \addplot[darkbrown, solid, line width=0.15cm] coordinates { (1,0.6) (1,0.1) };
        \addplot[blue, decorate, decoration={random steps,segment length=2pt,amplitude=1pt}] coordinates { (0,1) (1,0.60) };
        \addplot[blue, decorate, decoration={random steps,segment length=2pt,amplitude=1pt}] coordinates { (0,1) (1,0.55) };
        \addplot[blue, decorate, decoration={random steps,segment length=2pt,amplitude=1pt}] coordinates { (0,1) (1,0.50) };
        \addplot[blue, decorate, decoration={random steps,segment length=2pt,amplitude=1pt}] coordinates { (0,1) (1,0.45) };
        \addplot[blue, decorate, decoration={random steps,segment length=2pt,amplitude=1pt}] coordinates { (0,1) (1,0.40) };
        \addplot[blue, decorate, decoration={random steps,segment length=2pt,amplitude=1pt}] coordinates { (0,1) (1,0.35) };
        \addplot[blue, decorate, decoration={random steps,segment length=2pt,amplitude=1pt}] coordinates { (0,1) (1,0.30) };
        \addplot[blue, decorate, decoration={random steps,segment length=2pt,amplitude=1pt}] coordinates { (0,1) (1,0.25) };
        \addplot[blue, decorate, decoration={random steps,segment length=2pt,amplitude=1pt}] coordinates { (0,1) (1,0.20) };
        \addplot[blue, decorate, decoration={random steps,segment length=2pt,amplitude=1pt}] coordinates { (0,1) (1,0.15) };
        \addplot[blue, decorate, decoration={random steps,segment length=2pt,amplitude=1pt}] coordinates { (0,1) (1,0.10) };
      \end{axis}
    \end{tikzpicture}
    \caption{\small With reduced variance, all members of the power
      mean family (brown shading) converge to the deterministic model
      while their lower bounds tighten.}
    \label{fig:lower-bound-transition}
  \end{subfigure}
  \caption{\small Tightness of lower bounds vs evaluation time
    prediction variance in the extended dropout family.}
  \label{fig:lower-bounds-abound}
\end{figure}

Suppose we pick a base model from the power mean family and have a
continuum of subvariants with gradually reduced variance in their
predictions but the same expectation.
Clearly, for each of them we can derive a lower bound the same way as
we did for the power mean family.
And as we showed above, the lower bounds will \emph{tend to} increase
as the variance of the predictions decreases (see
Fig.\,\ref{fig:lower-bound-bounds}).
They do not strictly increase, only tend to, due to how the Jensen gap
is bounded from above and below and also due to the $\mathcal{O}$ term
of Eq.\,\ref{eq:jensen-gap-approximation}.
Nonetheless, as we approach determinism the lower bound is forced into
increasingly tighter ranges with strictly monotonically increasing
bounds around it, thus we can always reduce the variance such that
there is no overlap between the ranges and we get a guaranteed
improvement on the lower bound.
This effect reaches its apex at the deterministic model whose lower bound
is both exact and higher than any other model's.
Fig.\,\ref{fig:lower-bound-transition} illustrates that regardless of
the choice of base model, reducing the prediction variance will
eventually transform it into the same deterministic model.

\subsection{The extended power mean family: controlling the tightness of the bound}

Intuitively, in the absence of other sources of stochasticity the
dropout rate controls the variance of the predictions and if it is
low, the lower bound can be pretty snug.
However, there are two problems.

First, decreasing the dropout rate does not necessarily keep the
expectation of the predictions the same.
We offer no solution to this bias issue, but refer the reader to
previous studies of dropout's approximation properties such as
\citep{baldi2013understanding} and our subsequent empirical results.

Second, reducing the dropout rate would trade off generalisation for
tighter bounds.
But doing so only at evaluation time leaves the training time
regularisation effect intact, and can be seen as picking another model
whose lower bound tends to be higher than that of the base model.
Having thus extended the dropout family further, we can now tweak both
$\alpha$ and dropout rates at evaluation time.

Depending on the severity of the introduced bias compared to the
benefits of having a tighter lower bound, the optimal variance may lie
anywhere between the deterministic and the base model.
We show experimentally that across a number of datasets the benefits
of tighter bounds matter more, and observe monotonic improvement in
model fit as evaluation time dropout rates are decreased all the way
to full determinism.
The experiment was conducted as follows.
On an already trained model, the dropout rate was multiplied by
$\lambda \in [0,1]$.
\begin{table}
  \small
  \centering
  \caption{\small \ptb training XEs with various dropout rate
    multipliers between deterministic and GMC. Observe the monotonic
    improvement in training fit when reducing the dropout rate
    \emph{at evaluation only}.}
  \label{tab:ptb-dropout-rate-multiplier-training}
  \begin{tabular}{@{}rrrrrrrrrrr@{}}
    \toprule
    $\times$0.0 & $\times$0.1 & $\times$0.2 & $\times$0.3 & $\times$0.4 & $\times$0.5 & $\times$0.6 & $\times$0.7 & $\times$0.8  & $\times$0.9 & $\times$1.0\\
    \midrule
    2.731 & 2.738 & 2.746 & 2.755 & 2.766 & 2.777 & 2.791 & 2.807 & 2.826 & 2.849 & 2.878 \\
    \bottomrule
  \end{tabular}
\end{table}
As Table\,\ref{tab:ptb-dropout-rate-multiplier-training} shows, the
model fit as measured by cross entropy (XE) on the \emph{training set}
improves monotonically when reducing $\lambda$.
Results on other datasets and with other power mean models are very
similar.
We call the union of the reduced dropout rate subvariants of all power
mean family models the \textbf{extended dropout family} paramaterised
by $\alpha,\lambda$.

Therefore, we can say that dropout training optimises a deterministic
model subject to regularisation constraints, and deterministic
evaluation, widely believed to approximate MC evaluation, is the
closest match to the true objective at our disposal. It is not that
dropout evaluation has a deterministic approximation: dropout trains a
deterministic model first and foremost and a continuum of stochastic
ones to various extents.

In summary, we described dropout training as optimising a common lower
bound for a family of models.
Since this lower bound is the same for all models in the family, we
can nominate any of them at evaluation time.
However, the tightness of the bound varies, which affects model fit.
Having trained a model with dropout, the best fit is achieved by the
deterministic model with no dropout.
This result isolates the regularisation effects from the biases of the
lower bound and the dropout family.

\section{Applying dropout}
\label{sec:applying-dropout}

We investigate how members of the extended dropout model family
perform in terms of generalisation.
We follow the experimental setup of \cite{melis2017state} and base our
work on their best performing model variant for each dataset.
Unless explicitly stated, no retraining was performed and their model
weights reused.
In the experiments with the tuning objective, we follow their
experimental setup, using Google Vizier \citep{golovin2017google}, a
black-box hyperparameter tuner based on batched Gaussian Process
Bandits.

\begin{table}
  \small
  \centering
  \caption{\small Validation XEs on some datasets varying the power
    $\alpha$ and the dropout rate mulitiplier $\lambda$. Deterministic
    dropout is not the best evaluation method for the language
    modelling datasets due to a simple smoothing effect.}
  \label{tab:eval-methods-validation}
  \begin{tabular}{@{}lrrrrrrrrrr@{}}
    \toprule
    & & \multicolumn{3}{c}{Geometric ($\alpha=0$)} & \multicolumn{3}{c}{Power $\alpha=0.5$} & \multicolumn{3}{c}{Arithmetic ($\alpha=1$)} \\
    \cmidrule(lr){3-5} \cmidrule(lr){6-8} \cmidrule(l){9-11}
    Dataset & DET & $\times$0.8 & $\times$0.9 & $\times$1.0 & $\times$0.8 & $\times$0.9 & $\times$1.0 & $\times$0.8 & $\times$0.9 & $\times$1.0 \\
    \midrule
    MNIST      & \textbf{0.070} & 0.087 & 0.087 & 0.088 & 0.92 & 0.93 & 0.93 & 0.100 & 0.100 & 0.100 \\
    Enwik8     & 0.886 & 0.879 & 0.878 & 0.881 & 0.877 & 0.877 & 0.877 & 0.875 & 0.875 & \textbf{0.875} \\
    PTB        & 4.110 & 4.090 & 4.090 & 4.093 & 4.072 & 4.070 & 4.073 & \textbf{4.061} & 4.064 & 4.080 \\
    Wikitext-2 & 4.236 & 4.229 & 4.231 & 4.235 & 4.025 & 4.026 & 4.208 & \textbf{4.203} & 4.212 & 4.228 \\
    \bottomrule
  \end{tabular}
\end{table}

See Table\,\ref{tab:eval-methods-validation} for results of image
classification on MNIST, character based language modelling on Enwik8,
word based language modelling on PTB and Wikitext-2.
On MNIST, deterministic dropout is the best in terms of cross entropy,
which matches our theoretical predictions.
In contrast, on language modelling arithmetic averaging produces the
best results, which necessitates further analysis.

\begin{table}
  \small
  \centering
  \caption{\small PTB training and validation XEs for AMC at $\lambda
    \in \{0,0.8,1\}$ per word frequency. Note how DET dominates AMC on
    the training set, but AMC is better for rare words in the
    validation set.}
  \label{tab:xe-vs-frequency}
  \begin{tabular}{@{}lrrrrrrr@{}}
    \toprule
    & num     & \multicolumn{3}{c}{training} & \multicolumn{3}{c}{validation} \\
    \cmidrule(lr){3-5}              \cmidrule(l){6-8}
    frequency & targets &  DET & $\times$0.8 & AMC & DET & $\times$0.8 & AMC \\
    \midrule
    25000$<$ & 13580 & 1.40 & 1.50 & 1.56 & 1.58 & 1.64 & 1.68 \\
    5000$<$  & 26658 & 1.65 & 1.75 & 1.81 & 1.93 & 1.98 & 2.02 \\
    500$<$   & 44702 & 2.19 & 2.30 & 2.36 & 2.58 & 2.63 & 2.66 \\
    $<$500   & 29058 & 4.07 & 4.19 & 4.29 & 6.49 & 6.39 & 6.39 \\
    $<$100   & 14222 & 4.24 & 4.38 & 4.49 & 7.81 & 7.64 & 7.61 \\
    $<$20    &  5008 & 4.00 & 4.19 & 4.33 & 9.20 & 9.01 & 8.97 \\
    \bottomrule
  \end{tabular}
\end{table}

We suspected that the particularly severe form of class imbalance
exhibited by the power-law word distribution \citep{zipf1935psycho}
might play a role.
To verify this, we contrasted training and validation XEs on PTB for
words grouped by frequency (see Table\,\ref{tab:xe-vs-frequency}).
On the training set, the gap between deterministic dropout and AMC
is wider for low frequency words.
On the validation set, AMC is worse for frequent words but better for
rare words.
The $\times$0.8 dropout multiplier just finds a reasonable compromise.

\subsection{Softmax temperature}

The observed effect is consistent with smoothing, thus we posit that
the reason MNIST results are worse with AMC is that the marginal
distributions of labels in the training and test set are identical by
construction and further smoothing is unnecessary.
On the other hand, PTB and Wikitext-2 benefit from AMC's smoothing
because the penalty for underestimating low probabilities is harsh,
hence the large improvement on rare words.
The character based Enwik8 dataset lies somewhere in between: the
training and test distributions are better matched and there are no
very low probability characters.

\begin{table}
  \small
  \centering
  \caption{\small Validation and test perplexities on PTB and
    Wikitext-2 with various evaluation strategies and default or
    optimal validation softmax temperatures. Our baseline results
    correspond to DET at temperature 1. Note that AMC does not benefit
    from setting the optimal softmax temperature (``opt''), while DET
    is improved by it almost to the point of matching AMC which
    supports the smoothing hypothesis.}
  \label{tab:ptb-multiplier-temperature-validation}
  \begin{tabular}{@{}lrrrrrrrrrrrr@{}}
    \toprule
    & & & & \multicolumn{3}{c}{Geometric ($\alpha=0$)} & \multicolumn{3}{c}{Power $\alpha=0.5$} & \multicolumn{3}{c}{Arithmetic ($\alpha=1$)} \\
    \cmidrule(lr){5-7} \cmidrule(lr){8-10} \cmidrule(l){11-13}
    & Dataset & Temp & DET   & $\times$0.8  & $\times$0.9  & $\times$1.0 & $\times$0.8  & $\times$0.9  & $\times$1.0 & $\times$0.8  & $\times$0.9  & $\times$1.0 \\
    \midrule
    \parbox[t]{2mm}{\multirow{4}{*}{\rotatebox[origin=c]{90}{Validation~~}}}
    & \multirow{2}{*}{WT-2} &
    1 & \nlltoppl{4.236} & \nlltoppl{4.229} & \nlltoppl{4.231} & \nlltoppl{4.235} & \nlltoppl{4.205} & \nlltoppl{4.207} & \nlltoppl{4.208} & \nlltopplbold{4.203} & \nlltoppl{4.212} & \nlltoppl{4.228} \\
    & & opt & \nlltopplbold{4.211} & \nlltopplbold{4.212} & \nlltoppl{4.215} & \nlltoppl{4.220} & \nlltopplbold{4.205} & \nlltoppl{4.206} & \nlltoppl{4.208} & \nlltoppl{4.203} & \nlltoppl{4.210} & \nlltoppl{4.221} \\
    \cmidrule(l){2-13}
    & \multirow{2}{*}{PTB} & 1 & \nlltoppl{4.110} & \nlltoppl{4.088} & \nlltoppl{4.089} & \nlltoppl{4.089} & \nlltoppl{4.062} & \nlltoppl{4.059} & \nlltoppl{4.060} & \nlltoppl{4.048} & \nlltoppl{4.051} & \nlltoppl{4.069} \\
    & & opt & \nlltopplbold{4.051} & \nlltopplbold{4.052} & \nlltoppl{4.058} & \nlltoppl{4.065} & \nlltopplbold{4.045} & \nlltoppl{4.049} & \nlltoppl{4.057} & \nlltopplbold{4.044} & \nlltoppl{4.052} & \nlltoppl{4.067} \\
    \midrule
    \parbox[t]{2mm}{\multirow{4}{*}{\rotatebox[origin=c]{90}{Test~~}}}
    & \multirow{2}{*}{WT-2} &
          1 & \nlltoppl{4.188} & \nlltoppl{4.179} & \nlltoppl{4.180} & \nlltoppl{4.183} & \nlltopplbold{4.155} & \nlltoppl{4.157} & \nlltoppl{4.162} & \nlltopplbold{4.154} & \nlltoppl{4.166} & \nlltoppl{4.182} \\
    & & opt & \nlltopplbold{4.166} & \nlltopplbold{4.169} & \nlltoppl{4.171} & \nlltoppl{4.173} & \nlltoppl{4.155} & \nlltoppl{4.156} & \nlltoppl{4.162} & \nlltoppl{4.154} & \nlltoppl{4.162} & \nlltoppl{4.173} \\
          \cmidrule(l){2-13}
    & \multirow{2}{*}{PTB} & 1 & \nlltoppl{4.071} & \nlltoppl{4.049} & \nlltoppl{4.050} & \nlltoppl{4.050} & \nlltoppl{4.025} & \nlltoppl{4.022} & \nlltoppl{4.024} & \nlltopplbold{4.012} & \nlltoppl{4.016} & \nlltoppl{4.034} \\
    & & opt & \nlltopplbold{4.026} & \nlltopplbold{4.025} & \nlltoppl{4.027} & \nlltoppl{4.035} & \nlltoppl{4.020} & \nlltopplbold{4.020} & \nlltoppl{4.026} & \nlltoppl{4.012} & \nlltoppl{4.016} & \nlltoppl{4.031} \\
    \bottomrule
  \end{tabular}
\end{table}

To test the hypothesis that AMC's advantage lies in smoothing, we
tested how performing smoothing by other means affects the results.
In this experiment, on a trained model the temperature of the final
softmax was optimised on the validation set and the model was applied
with the optimal temperature to the validation and test sets.
Our experimental results in Table
\ref{tab:ptb-multiplier-temperature-validation} support the hypotheses
that AMC smooths the predicted distribution as increasing the
temperature improves DET and GMC considerably but not AMC.
In fact, the optimal temperature for AMC with $\lambda=1$ was slightly
lower than 1, which corresponds to sharpening, not smoothing.

Tuning the evaluation time softmax temperature is similar to label
smoothing \citep{pereyra2017regularizing}, the main difference being
that our method does not affect training.
While this is convenient, for tuning model hyperparameters, ideally we
would determine the optimal evaluation parameters $\alpha$, $\lambda$
and the temperature for the calculation of the validation score for
each set of hyperparameters tried, but this would be prohibitively
expensive.
Since deterministic evaluation coupled with the optimal temperature is
very close to the best performing AMC model, it serves as a good proxy
for the ideal tuning objective.
The optimal temperature can be approximately determinined using a
linear search on a subset of the validation data which is orders of
magnitude faster than MC dropout.
In our experiments, hyperparameter tuning with validation scores
computed at the optimal softmax temperature did improve results,
albeit very slightly (about half a perplexity point).
Thus we can conclude that deterministic dropout is already a
reasonable proxy for which to optimise.

\subsection{Results}

We have improved the best test result of \cite{melis2017state} from
\nlltopplbold{4.065} to \nlltopplbold{4.020} on PTB, and from
\nlltopplbold{4.188} to \nlltopplbold{4.154} on
Wikitext-2 using their model weights, only tuning the evaluation
parameters $\alpha$, $\lambda$ and the softmax temperature on the
validation set.
By retuning the hyperparameters of the PTB model with optimal
temperature deterministic evaluation, we improved to
\nlltopplbold{4.012} on PTB.
For lack of resources, we did not retune for Wikitext-2.
For comparison, the state of the art in language modelling without
resorting to dynamic evaluation or a continuous cache pointer is
Mixture of Softmaxes \citep{yang2017breaking} with 54.44 and 61.45 on
PTB and Wikitext-2, respectively.
At present, it is unclear whether the benefits of their approach and
ours combine.

In summary, we looked at how different models and evaluation methods
rank in terms of generalisation.
Across a number of tasks and datasets the ranking differed from what
was observed on the training set.
We found that AMC smooths the distribution of the prediction
probabilities and we achieved a similar effect without resorting to
expensive sampling simply by adjusting the temperature of the final
softmax.
Finally, we brought the tuning objective more in line with the
improved evaluation by automatically determining the optimal softmax
temperature when evaluating on the validation set which further
improved results.

\section{Implications}

The construction of a conditional model family with a common lower
bound on their objectives is applicable to other latent variable
models with similar structure and inference method.
This lower bound admits ambiguity as to what model is being fit to the
data, which in turn allows for picking any such model at evaluation
time.
However, the tightness of the bound and the quality of the fit varies.
For dropout, the deterministic model has the best fit even though the
training objective is highly stochastic, but this result hinges on the
approximation properties of deterministic dropout and will not carry
over to other probabilistic models in general.
In particular, standard VAEs \citep{kingma2013auto} with their lower
bound being very similar in construction to Eq.\,\ref{eq:loss} cannot
quite collapse to a deterministic model else they suffer an infinite
KL penalty.
Still, the lower bound being looser on the tails of $q$ is related to
problem of underestimating posterior uncertainty
\citep{turner2011two}.

In related works, expectation-linear dropout
\citep{DBLP:journals/corr/MaGHYDH16} and fraternal dropout
\citep{zolna2017fraternal} both try to reduce the ``inference gap'':
the mismatch between the training objective and deterministic
evaluation.
The gains reported in those works might be explained by reducing the
bias of deterministic evaluation and also by encouraging small
variance in the predictions and thus getting tighter bounds.
Another recent work, activation regularisation
\citep{merity2017regularizing}, could be thought of as a mechanism to
reduce the variance of predictions to a similar effect.
In the context of language modelling, the connection between noise and
smoothing was established by \cite{xie2017data}.
Our improved understanding further emphasises that connection, and at
the same time challenges the way we think about dropout.

\subsubsection*{Acknowledgments}

We would like to thank Laura Rimell, Aida Nematzadeh, and Andriy
Mnih for their valuable feedback.

\bibliography{paper}
\bibliographystyle{iclr2019_conference}

\clearpage
\begin{appendices}

\section{Variational dropout with non-shared masks}
\label{sec:non-shared-mask}

If $q$ and $p$ are redefined for the non-shared setting to be products
of identical and independent, per time step factors, neither term of
the variational objective requires rethinking: the MC approximation
still works since $q$ is easy to sample from, while the KL term
becomes a sum of componentwise KL divergences and can still be
implemented as weight decay.
Consequently, both shared and non-shared masks fit into the
variational framework.
For a detailed derivation see
Appendix\,\ref{sec:non-shared-mask-derivation}.

In related works, \cite{pachitariu2013regularization} in their
investigation of regularisation of standard RNN based language models
dismiss applying dropout to recurrent connections ``\textit{to avoid
  introducing instabilities into the recurrent part of the LMs}''.
\cite{bayer2013fast} echo this claim about RNNs, which is then cited
by \cite{zaremba2014recurrent}, but their work is based on LSTMs not
standard RNNs.
Finally, \cite{gal2016theoretically} cite all of the above but also
work with LSTMs.
Their results indicate a large, about 15 perplexity point
advantage to shared mask dropout for language modelling on the Penn
Treebank (\ptb) corpus (see Fig.\,2 in their paper).

Our experimental results obtained with careful and extensive
hyperparameter tuning, listed in
Table\,\ref{tab:ptb-shared-vs-nonshared}, indicate only a small
difference between the two which is in agreement with the empirical
study of \cite{semeniuta2016recurrent}.

\begin{table}[h]
  \small
  \centering
  \caption{\small Validation and test set perplexities on \ptb with
    shared (S) or non-shared (NS) dropout masks for a small, 1 layer
    and a large, 4 layer LSTM with 10 and 24 million weights,
    respectively. Non-shared masks perform nearly as well as shared
    masks and as we have seen neither is ``more variational'' than the
    other.}
  \label{tab:ptb-shared-vs-nonshared}
  \begin{tabular}{@{}lrrrr@{}}
    \toprule
    dataset & \multicolumn{2}{c}{10M} & \multicolumn{2}{c}{24M} \\
    \cmidrule(lr){2-3}\cmidrule(l){4-5}
    & \multicolumn{1}{c}{S} & \multicolumn{1}{c}{NS} & \multicolumn{1}{c}{S} & NS~ \\
    \midrule
    validation & \nlltoppl{4.084} & \nlltoppl{4.098} & \nlltoppl{4.051} & \nlltoppl{4.066} \\
    test & \nlltoppl{4.052} & \nlltoppl{4.071} & \nlltoppl{4.026} & \nlltoppl{4.041} \\
    \bottomrule
  \end{tabular}
\end{table}

In any case, non-shared masks, in addition to being variational, are
also surprisingly competitive with shared masks for LSTMs (we make no
claims about standard RNNs).
We also tested whether embedding dropout (in which dropout is applied
to entire vectors in the input embedding lookup table) proposed by
\cite{gal2016theoretically} improves results, and find that embedding
dropout does not offer any improvement on top of input dropout.


\section{Derivation of variational dropout with non-shared masks}
\label{sec:non-shared-mask-derivation}

In this section, we formulate naive (i.e. non-shared mask) dropout in
the variational setting.
In contrast to the shared mask case, where $\w$ was a single set of
weights, here $\w^{1:T}$ (or $\w$, for short) has a set of weights for
each time step that differ in their dropout masks.
The variational posterior $q(\w)$ and the prior $p(\w)$ are both
products of identical distributions over time:
\begin{align*}
    q(\w^{1:T}) &= \prod_{t=1}^T q'(\w^t)
                    = \prod_{t=1}^T \big[ p \N(\w^t|0, \sigma^2) +
                                      (1-p) \N(\w^t| \Theta, \sigma^2)\big]\\
    p(\w^{1:T}) &= \prod_{t=1}^T p'(\w^t)
                    = \prod_{t=1}^T \N(\w^t|0, \sigma_p^2)
\end{align*}
An unbiased approximation to the integrals in Eq.\,\ref{eq:loss} is
based on a single, easy to obtain sample $\hat{\w} \sim q(\w)$:
\begin{align*}
    \int q(\w) \ln p(y|x,\w) d\w \approx \ln p(y|x,\hat{\w})
\end{align*}
Showing that the KL term can still be approximated with weight decay
with non-shared masks is not much more involved.
Both distributions are products of densities over independent random
variables, so the componentwise KL divergencies sum.
In particular:
\begin{align*}
    \KL(q(\w)||p(\w)) &=
    \int \bigg(\prod_{i=1}^T q'(\w^i)\bigg)
      \ln \frac{\prod_{t=1}^T q'(\w^t)}{\prod_{t=1}^T p'(\w^t)} d\w \\
    &= \int \bigg(\prod_{i=1}^T q'(\w^i)\bigg) 
      \sum_{t=1}^T \ln \frac{q'(\w^t)}{p'(\w^t)} d\w \\
    &= \sum_{t=1}^T \int \bigg(\prod_{i=1}^T q'(\w^i)\bigg)
      \ln \frac{q'(\w^t)}{p'(\w^t)} d\w \\
    &= \sum_{t=1}^T \int \bigg[\prod_{i=1, i\neq t}^T q'(\w^i)\bigg] \bigg[q'(\w^t)
      \ln \frac{q'(\w^t)}{p'(\w^t)}\bigg] d\w \\
    &= \sum_{t=1}^T \bigg[\int \prod_{i=1, i\neq t}^T q'(\w^i) d\w^{\setminus t}\bigg]
    \bigg[\int q'(\w) \ln \frac{q'(\w)}{p'(\w)} d\w\bigg] \\
    &= T \cdot \KL(q'(\w)||p'(\w))
\end{align*}
We partitioned the variables into two mutually exclusive sets $w^t$
and its complement $w^{\setminus t}$, and split the multiple integral
using Fubini's theorem (or, equivalently, using the expectation of
independent random variables rule). 
After the split, the first integral is trivially $1$ and the second
has no dependence on $T$.

What we end up with is a sum of identical KL terms of the same
distributions as in the shared mask case, so the full KL can be
approximated with weight decay.

\section{Derivation of the MAP lower bound for the arithmetic model}
\label{sec:map-objective-derivation}
We can rewrite the posterior as:
\begin{align*}
    p(\Theta|X,Y) &= \frac{p(X,Y|\Theta)p(\Theta)}{p(X,Y)}\\
    &\propto p(X,Y|\Theta)p(\Theta)\\
    &= \int p(Y|X,\w,\Theta) p(\w|\Theta,X) p(\Theta|X) p(X) d\w\\
    &\propto \int p(Y|X,\w) p(\w|\Theta) p(\Theta) d\w
\end{align*}
Moving to the log domain and using Jensen's inequality allows us to
construct a lower bound that is a sum of per data point terms (i.e.
something that can be conveniently optimised):
\begin{align*}
    \ln p(\Theta|X,Y) &= \ln \int p(Y|X,\w) p(\w|\Theta) p(\Theta) d\w - C_{\textit{MAP}}\\
    &= \ln \int p(\w|\Theta) \prod_{i=1}^N p(y_i|x_i,\w) d\w + \ln p(\Theta) - C_{\textit{MAP}}\\
    &\geqslant \int p(\w|\Theta) \ln \prod_{i=1}^N p(y_i|x_i,\w) d\w + \ln p(\Theta) - C_{\textit{MAP}}\\
    &= \sum_{i=1}^N \int p(\w|\Theta) \ln p(y_i|x_i,\w) d\w + \ln p(\Theta) - C_{\textit{MAP}}
\end{align*}

\section{Derivation of the MAP lower bound for the geometric model}
\label{sec:geometric-objective-derivation}

From Eq.\,\ref{eq:renormalised-geometric-model} recall that:
\begin{align*}
    p(y|x, \Theta) &=
    \frac{\exp\left(\E_{\hat{\w} \sim p(\w|\Theta)} \ln p(y|x,\hat{\w})\right)}{Z(x,\Theta)}
\end{align*}
The normalisation constant $Z$ is at most $1$, due to the geometric
mean being bounded from above by the arithmetic mean on a per class
$c$ basis:
\begin{align*}
  Z(x,\Theta) &= \sum_{c=1}^C exp\big(\E_{\hat{\w} \sim p(\w|\Theta)} \ln p(c|x,\hat{\w})\big)\\
    &\leqslant \sum_{c=1}^C \E_{\hat{\w} \sim p(\w|\Theta)} p(c|x,\hat{\w})\\
    &= \E_{\hat{\w} \sim p(\w|\Theta)} \sum_{c=1}^C p(c|x,\hat{\w}) = 1
\end{align*}
Since this a conditional model, we can rewrite the posterior as:
\begin{align*}
    p(\Theta|X,Y) &= \frac{p(X,Y|\Theta)p(\Theta)}{p(X,Y)}\\
    &\propto p(X,Y|\Theta)p(\Theta)\\
    &= p(Y|X,\Theta) p(X|\Theta) p(\Theta)\\
    &\propto p(Y|X,\Theta) p(\Theta)
\end{align*}
$p(X|\Theta)$ is dropped in the last step as it is constant.
Moving to the log domain once again:
\begin{align*}
  \ln p(\Theta|X,Y) &= \ln p(Y|X,\Theta) + \ln p(\Theta) - C_{MAP}\\
  &= \ln \prod_{i=1}^N p(y_i|x_i,\Theta) + \ln p(\Theta) - C_{MAP}\\
  &= \sum_{i=1}^N \bigg[\E_{\hat{\w} \sim p(\w|\Theta)} \ln p(y_i|x_i,\hat{\w}) - \ln(Z(x_i,\Theta))\bigg] + \ln p(\Theta) - C_{MAP}\\
  &\geqslant \sum_{i=1}^N \E_{\hat{\w} \sim p(\w|\Theta)} \ln p(y_i|x_i,\hat{\w}) + \ln p(\Theta) - C_{MAP}\\
  &= \sum_{i=1}^N \int p(\w|\Theta) \ln p(y_i|x_i,\w) d\w + \ln p(\Theta) - C_{MAP}
\end{align*}
where the lower bound arises due to $\forall i\colon Z(x_i,\Theta)\leqslant1$.

\section{Derivation of the MAP lower bound for the power mean family}
\label{sec:power-mean-objective-derivation}

In §\ref{sec:power-mean-family} we proved that $\forall i\colon
Z(x_i,\Theta)\leqslant 1$.
Starting from $p(\Theta|X,Y) \propto p(Y|X,\Theta) p(\Theta)$ just
like in the geometric case, we derive a lower bound in the log domain:
\begin{align*}
  \ln p(\Theta|X,Y) &= \ln p(Y|X,\Theta) + \ln p(\Theta) - C_{\textit{MAP}}\\
  &= \ln \prod_{i=1}^N p(y_i|x_i,\Theta) + \ln p(\Theta) - C_{\textit{MAP}}\\
  &= \sum_{i=1}^N \bigg[\ln \sqrt[\alpha]{\E_{\hat{\w} \sim p(\w|\Theta)} p(y_i|x_i,\hat{\w})^\alpha} - \ln(Z(x_i,\Theta))\bigg] + \ln p(\Theta) - C_{\textit{MAP}}\\
  &\geqslant \sum_{i=1}^N \ln \sqrt[\alpha]{\E_{\hat{\w} \sim p(\w|\Theta)} p(y_i|x_i,\hat{\w})^\alpha} + \ln p(\Theta) - C_{\textit{MAP}}\\
  &= \sum_{i=1}^N \frac{1}{\alpha} \ln \bigg(\E_{\hat{\w} \sim p(\w|\Theta)} p(y_i|x_i,\hat{\w})^\alpha\bigg) + \ln p(\Theta) - C_{\textit{MAP}}\\
  &\geqslant \sum_{i=1}^N \frac{1}{\alpha} \E_{\hat{\w} \sim p(\w|\Theta)} \ln p(y_i|x_i,\hat{\w})^\alpha + \ln p(\Theta) - C_{\textit{MAP}}\\
  &= \sum_{i=1}^N \E_{\hat{\w} \sim p(\w|\Theta)} \ln p(y_i|x_i,\hat{\w}) + \ln p(\Theta) - C_{\textit{MAP}}\\
  &= \sum_{i=1}^N \int p(\w|\Theta) \ln p(y_i|x_i,\w) d\w + \ln p(\Theta) - C_{\textit{MAP}}
\end{align*}

\end{appendices}

\end{document}